\pdfoutput=1

\documentclass[11pt]{article}

\usepackage{acl}

\usepackage{times}
\usepackage{latexsym}
\usepackage{graphicx}
\usepackage{multicol}
\usepackage{booktabs}
\usepackage{amsfonts}
\usepackage{amsmath}
\usepackage{comment}
\usepackage[T1]{fontenc}
\usepackage{adjustbox}

\usepackage[utf8]{inputenc}

\usepackage{microtype}

\title{Coalescing Global and Local Information for Procedural Text Understanding}

\author{
Kaixin Ma$^{\dagger}$,
Filip Ilievski$^{\S}$,
Jonathan Francis$^{\dagger\P}$,\\ 
\textbf{Eric Nyberg$^{\dagger}$,
Alessandro Oltramari$^{\P}$}\\
  $^{\dagger}$Language Technologies Institute, Carnegie Mellon University\\
  $^{\S}$Information Sciences Institute, University of Southern California\\
  $^{\P}$Human-Machine Collaboration, Bosch Research Pittsburgh\\
\small\{kaixinm, jmf1, ehn\}@cs.cmu.edu, ilievski@isi.edu, alessandro.oltramari@us.bosch.com
\vspace{2mm}
}

\begin{document}
\maketitle
\begin{abstract}
Procedural text understanding is a challenging language reasoning task that requires models to track entity states across the development of a narrative. A complete procedural understanding solution should combine three core aspects: 
local and global views of the inputs, and global view of outputs. Prior methods considered a subset of these aspects, resulting in either low precision or low recall. In this paper, we propose \textbf{C}oalescing \textbf{G}lobal and \textbf{L}ocal \textbf{I}nformation (CGLI), a new model that builds \textit{entity-} and \textit{timestep-aware} input representations (local input) considering the \textit{whole} context (global input), and we \textit{jointly} model the entity states with a structured prediction objective (global output). Thus, CGLI simultaneously optimizes for both precision and recall. We extend CGLI with additional output layers and integrate it into a story reasoning framework. Extensive experiments on a popular procedural text understanding dataset show that our model achieves state-of-the-art results; experiments on a story reasoning benchmark show the positive impact of our model on downstream reasoning. We release our code here: \url{https://github.com/Mayer123/CGLI}
\end{abstract}

\section{Introduction}

\begin{figure}[!t]
    \centering
    \includegraphics[width=\columnwidth]{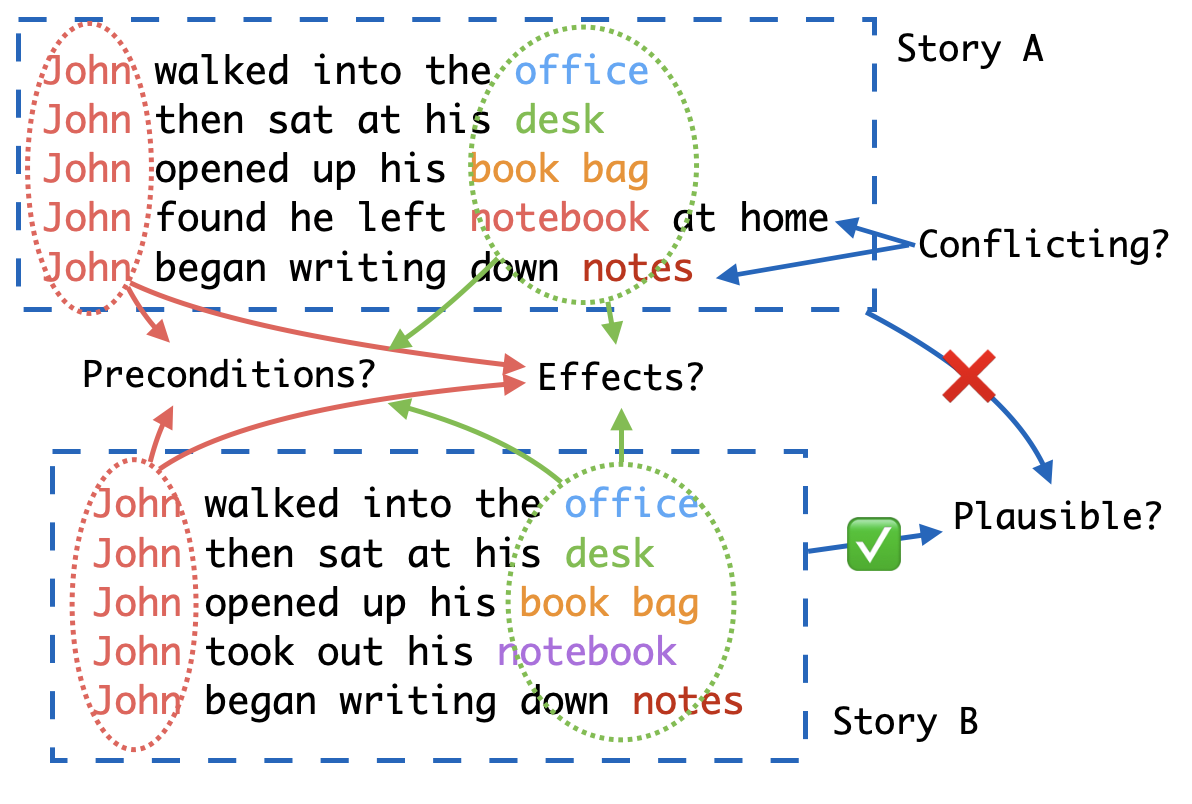}
    \caption{An example story of understanding task. Given two stories, the task is to judge which story is more plausible, find the conflicting sentence pair in the implausible story, and predict entity states at every step.}
    \label{fig:exp}
\end{figure}

\begin{figure*}[!ht]
    \centering
    \includegraphics[width=1.0\linewidth]{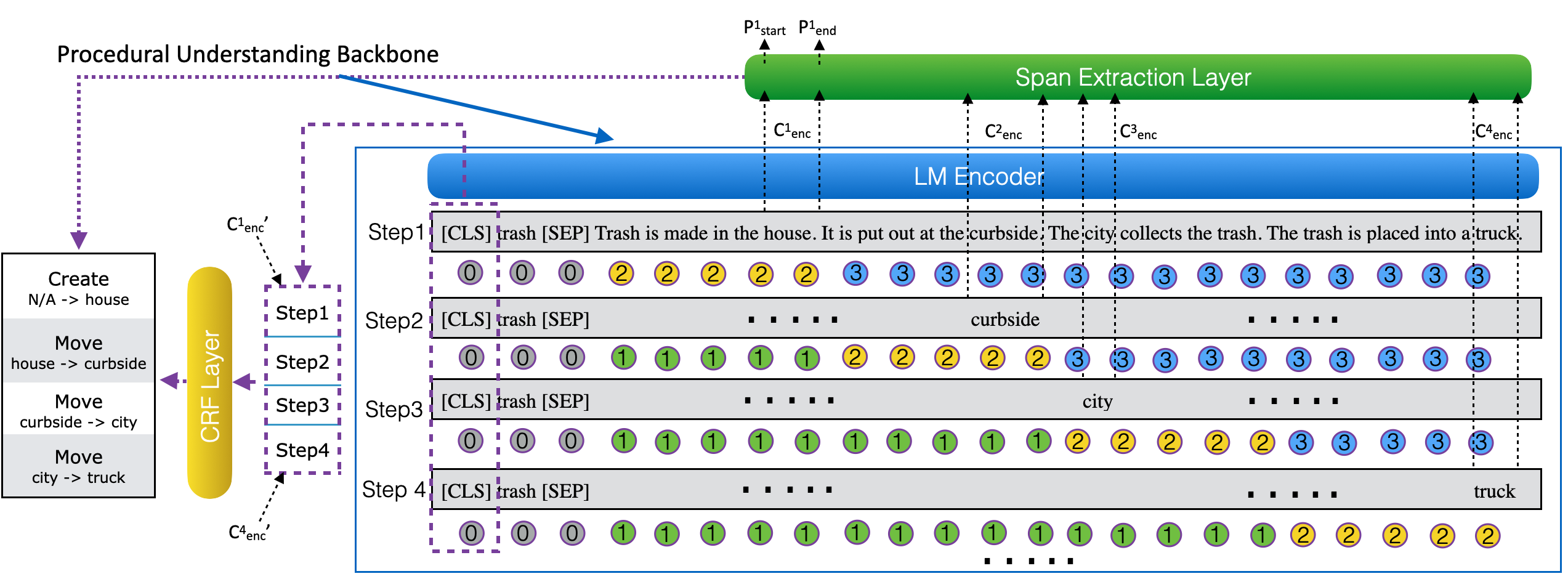}
    \caption{An illustration of CGLI. At every step, the LM encodes the full paragraph with different timestep ids (colored circles with numbers). The span extraction layer yields a location span for every entity at every step and this span sequence is combined with action sequence produced by a CRF layer to form the final predictions.}
    \label{fig:model}
\end{figure*}

Understanding the causal links of events in procedures is a key aspect of intelligence, facilitating human narration and dialogue. For instance, understanding why story B is plausible and why story A is not (\autoref{fig:exp}) requires procedural understanding of the causes of John leaving his notebook at home, as opposed to him taking out his notebook from his bag: writing in a notebook is counterfactual in the former case, and intuitive in the latter.
Understanding stories requires procedural models that can reason consistently about event implications, and do so at different granularities. For a model to decide whether a story is plausible, it has to track the entity states over time, understand the effects of the described actions (green arrows), and consider the preconditions for a given action (pink arrows). Meanwhile, the model must reconcile the causes and effects of all events described in the story, to provide a globally consistent interpretation. 

While procedural reasoning research reports steady progress in recent years~\cite{rajaby-faghihi-kordjamshidi-2021-time,gupta-durrett-2019-effective,zhang2021koala}, story understanding and procedural reasoning have rarely been considered together~\cite{storks-etal-2021-tiered-reasoning}.
Works have attended only to complementary aspects of the procedural reasoning problem, e.g., \citet{gupta-durrett-2019-effective} build entity-centric context representations and ignoring timestep-wise representation modeling; and \citet{rajaby-faghihi-kordjamshidi-2021-time} later proposed a timestep-specific model providing unique context encoding at every step to enable modeling flexibility. However, these methods predict independent step-wise entity states, thus compromising the dependency of outputs \textit{across} different steps---yielding high recall but low precision. 
Global-output methods~\cite{gupta-durrett-2019-tracking,zhang2021koala} explicitly leverage the strong dependency across steps by jointly modeling the entity actions from all steps, but these methods only have one context encoding for all entities and steps, thus providing sub-optimal input representations---yielding high precision but low recall.

In this paper, we propose \textbf{C}oalescing \textbf{G}lobal and \textbf{L}ocal \textbf{I}nformation (CGLI): a new model for procedural text understanding that makes global decisions in consideration of entity-, timestep-centric, and global views of the input. To do so, our model builds a separate input view for every entity, at every step, while providing the whole context. Meanwhile, CGLI represents the entity actions across steps jointly with a structured prediction objective, thus achieving high consistency between different steps. The contributions of this paper are:
    
\noindent \textbf{1. A novel procedural understanding method, CGLI}, which produces global outputs of narrative procedures based on a unified view of the input, combining both local (entity-centric, timestep-specific) and global (document-wide) views---thus optimizing precision and recall, simultaneously.

\noindent \textbf{2. A story understanding framework}, which builds upon our procedural understanding model, to enable story understanding with explicit and explainable understanding of event procedures, captured through entity precondition and effect states.

\noindent \textbf{3. An extensive evaluation} of CGLI against strong baselines on a procedural task, \texttt{ProPara}~\cite{dalvi-etal-2018-tracking}, and recent story understanding task, \texttt{TRIP}~\cite{storks-etal-2021-tiered-reasoning}. Our experiments show the positive impact of our method, through achieving state-of-the-art results, while ablation studies measure the impact of its individual components. 

\section{Task Definition}
\label{subsec:task1}
\textbf{Procedural text understanding.} The task input consists of an $n$-sentence paragraph $P = \{s_1, s_2, ... s_n\}$ , and $k$ entities $\{E_1, E_2, ... E_k\}$. The goal is to predict precondition state $S_{i,t}^p$ and effect state $S_{i,t}^e$, for every entity at every step, as well as the action $A_{i,t}$ performed by the entity at every step; $i \in \{1,2,..k\}$, $t \in \{1,2,...n\}$. The effect state at $t-1$ is the same as precondition state at step $t$, i.e., $ S_{i,t-1}^e = S_{i,t}^p$, hence $S_i$ is a sequence of length $n+1$. 
Following prior work~\cite{mishra2018tracking}, $A_{i,t} \in$ \{Create, Exist, Move, Destroy\}, $S_{i,t}^e \in $\{\textit{non-existence}, \textit{unknown-location}, \textit{location}\}, and for \textit{location}, a span of text in $P$ needs be identified for the prediction.
Action $A_{i,t}$ describes the entity state changes from precondition to effect, thus it can be inferred from the state sequence $S_i$, and vice versa---e.g., if $S_{i,1}^p =$ \textit{non-existence} and $S_{i,1}^e =$ \textit{location}, then $A_{i,1} =$ Create. 

\noindent \textbf{Procedural story understanding.}
The input to the procedural story understanding task consists of two parallel stories, $P_1$, $P_2 = \{s_1, s_2, ... s_n\}$, each consisting of $n$ sentences and differing only in one of the sentences. Following~\citet{storks-etal-2021-tiered-reasoning}, the task is to identify which story is more plausible, identify the conflicting pair of sentences ($s_{c1}$ and $s_{c2}$) in the implausible story, and list the preconditions $ S_{i}^e$ and effects $S_{i}^p$ of all entities at every step of a story. Here, multiple attributes are considered for precondition and effect states. Unlike in the procedural text understanding task, the story completion task does not require that the effect state at step $t-1$ should match the precondition state at step $t$, i.e., $S_{it-1}^e$ and $S_{it}^p$ are not necessarily equal. 


\section{CGLI: Coalescing Global and Local Information}
\label{sec:model}
In this section, we describe the input representation, the architecture, and the training details of our model, 
as illustrated in \autoref{fig:model}.



\noindent \textbf{Input representation.}
To allow greater modeling flexibility and enable span extraction for entity location-prediction,
we build a unique input representation for every entity at each step (\textit{local view}), and we provide it access to the entire context (\textit{global view}). Given an entity, we create a pseudo question $Q$ 'where is \{entity\}' (\textit{entity-aware}), and concatenate it with the full paragraph $P$, resulting in $C$ = [CLS] $Q$ [SEP] $P$ [SEP]. We map $C$ 
using the embedding layer of a language model (LM), resulting in $C_{emb}$. We then combine $C_{emb}$ with timestep embeddings to mark the current step of interest (\textit{timestep-aware}), following \cite{rajaby-faghihi-kordjamshidi-2021-time}. In particular, each input token is assigned a timestep ID where \{0=padding, 1=past, 2=current, 3=future\}, forming $T \in \mathbb{R}^{m}$, where $m$ is the number of tokens. The timestep sequence is projected through another embedding layer $Timestep \in \mathbb{R}^{4 \times d}$. The sum of $C_{emb}$ and $Timestep(T)$, denoted with $C_{emb}' \in \mathbb{R}^{d \times m}$, is then processed by the LM encoder layers, where $d$ is the hidden layer dimension of the LM encoder. Formally:\footnote{To model the precondition state of step 1, we also build an input sequence for step 0.}
\begin{align}
    C_{emb} & = \text{Embed}(C) \\
    C_{emb}' & = C_{emb} + Timestep(T) \\
    C_{enc} & = \text{LM Encoder}(C_{emb}')
\end{align}

\noindent \textbf{Location prediction.}
Given the LM encoded representation $C_{enc} \in \mathbb{R}^{d \times m}$, we extract the start and end indices of the location span: 
\begin{align}
    P_{Start} & = \text{Softmax}(W_s C_{enc})  \\
    P_{End} & = \text{Softmax}(W_e C_{enc}), 
\end{align}
where $W_s,W_e \in \mathbb{R}^{d}$. For unknown locations and non-existing states, we extract the [CLS] token as the span, analogous to how unanswerable questions are usually handled \cite{rajpurkar-etal-2018-know}.

\noindent \textbf{In-batch Conditional Random Field.}
For entity state/action modeling, we jointly predict the entity actions across all steps (\textit{global output}). We first group the encoded representation $C_{enc}^t$ of the same entity at different time steps $t$ in one batch chronologically, yield $C_{enc}^N \in \mathbb{R}^{d \times m \times (n+1)}$. Then we extract the [CLS] token embedding to represent the entity state of every step $C_{enc}^{N'} \in \mathbb{R}^{d \times (n+1)}$. We concatenate the entity state representation of every two consecutive steps to represent the actions between these two-state pairs. The result $D_{enc}^{N} \in \mathbb{R}^{2d \times n}$ is mapped to the emission scores $\phi \in \mathbb{R}^{a \times n}$, where $a$ is the number of possible actions. 
\begin{align}
    D_{enc}^t & = \text{Concat}(C_{enc}^{t'}, C_{enc}^{(t+1)'}) \\
    \phi & = W^T_a (tanh (W^T_d D_{enc}^{N}))
\end{align}
where  $W_d \in \mathbb{R}^{2d \times d}$, $W_a \in \mathbb{R}^{d \times a}$. The entity action sequence $A \in \mathbb{R}^{n}$ is modeled by a conditional random field (CRF):
{\small
\begin{align}
P(A|\phi, \psi) &  \propto \exp (\sum_{t=1}^{n} \phi_{t}(A_{t})+\psi(A_{t-1}, A_{t})),
\end{align}}%
with the CRF layer's  transition scores $\psi \in \mathbb{R}^{a \times a}$. \\
\noindent \textbf{Prior initialization.}
Previous methods~\cite{gupta-durrett-2019-tracking,zhang2021koala} initialize the CRF transition scores randomly and update them during training. This allows transition between any pair of actions. However, certain transitions between entity actions are nonsensical, e.g., an entity cannot be destroyed if it has not been created, and a destroyed entity cannot move. Learning such constraints may be possible if we have sufficient data, which is not the case for the tasks we are considering. Thus, we propose to directly impose commonsense constraints on the model's transition scores, because these conditions are universally true and can be used to reduce the model's search space. Specifically, we set an entity action transition score to \textit{-inf} if it has not been seen in the training data, otherwise we estimate the initial score of a transition based on its frequency in the training data: $\psi^{uv}$ = $log(\frac{Num(u, v)}{Num(u)})$,
where $\psi^{uv}$ is the log probability of transition from action $u$ to action $v$, $Num (u, v)$ is the transition count from $u$ to $v$ in data, $Num (u)$ is the count of $u$ in data. \\
\noindent \textbf{Training and inference.}
We jointly optimize the location and the entity action prediction losses during training:
\begin{align}
    \mathcal{L}_{loc} = - \frac{1}{n} \sum^{t=0}_n & (log(P_{Start}^{y_s^t}) + log(P_{End}^{y_e^t}))  \\
    \mathcal{L}_{action} & = -log(P(A|\phi, \psi)) \\
    \mathcal{L} = & \mathcal{L}_{loc} + \mathcal{L}_{action},
\end{align}

\noindent where $y_s^t$ and $y_e^t$ are the ground-truth start and end indices at step $t$. During inference, we use Viterbi decoding to produce the most likely entity action sequence and use the span extractor for the most likely location at every step. We combine the action sequence and location predictions to deterministically infer all precondition and effect states. 

\noindent \textbf{Data augmentation.}
Procedural text understanding requires dense annotation of entity states per step, making it challenging and expensive to collect large data. To address data sparsity, we propose a data augmentation method that could effectively leverage the unannotated paragraphs to enhance model's performance. In particular, we first train a model on the gold training set and then apply it to label the unannotated paragraphs, resulting a set of noisy examples. We then mix these examples with gold training data to train a second model. 






\begin{figure}[!t]
    \centering
    \includegraphics[width=\linewidth]{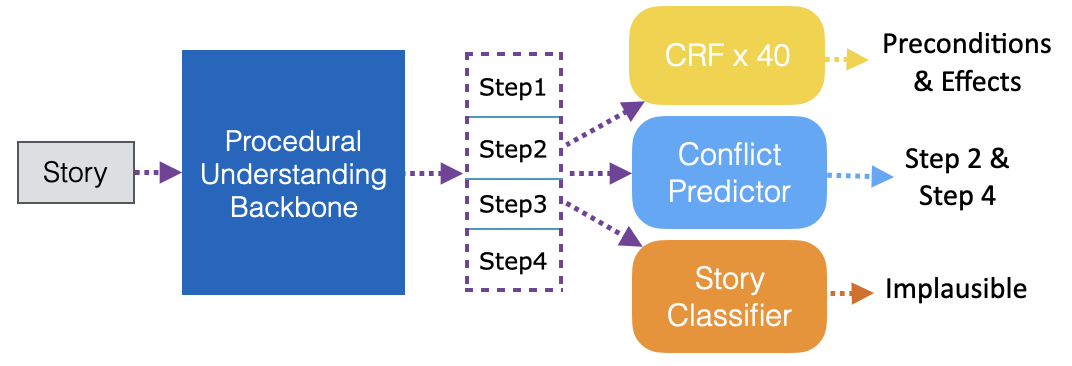}
    \caption{An illustration of integrating CGLI into a story understanding framework. The story is encoded in the same way as shown in \autoref{fig:model}, producing a sequence of step representations, i.e., a batch of [CLS] vectors. These vectors serve as input to different output layers to model the three task objectives: plausibility (orange), conflict sentence detection (blue), and entity state prediction (yellow).}
    \label{fig:model_story}
\end{figure}

\section{Story Understanding with CGLI}
\label{subsec:adapt}
We integrate CGLI into a story understanding framework with minimum modifications following the task definition, and the overall model is shown in \autoref{fig:model_story}. As the story understanding tasks do not require location extraction, we remove the span extraction layer, which makes the input representation of step 0 obsolete. Given that the continuity of effects to preconditions between consecutive steps does not hold in this task, we directly use $C_{enc}^{N'} \in \mathbb{R}^{d \times n}$ instead of $D_{enc}^{N} \in \mathbb{R}^{2d \times n}$ in the in-batch CRF. Given $B$ number of attributes for precondition and effect states, we apply an in-batch CRF module for each attribute. Specifically, we apply equations 7 and 8 for every attribute, yielding $2B$ such modules in total. 
To detect conflicting sentences, we concatenate every pair of sentence representations, and pass it through a linear layer to find the conflicting pair. For story classification, we take the mean of sentence representations for story representation, and pass it through a linear layer for binary classification. Formally, 
\begin{align}
   C_{confl} & = \text{vstack}(\text{Concat}(C_{enc}^{t'}, C_{enc}^{j'})) \\
   P_{confl} & = \text{Softmax}(W_{confl} C_{confl}) \\
   C_{plau} & = \text{Mean}(C_{enc}^{N'}) \\
   P_{plau} & = \text{Softmax}(W^T_{plau} C_{plau}), 
\end{align}
where $C_{confl} \in \mathbb{R}^{2d \times \frac{n(n-1)}{2}}$, $j \in \{{t+1},...n\}$, $W_{confl} \in \mathbb{R}^{2d}$, $C_{plau} \in \mathbb{R}^{d}$, $W_{plau} \in \mathbb{R}^{d \times 2}$. During training, we jointly optimize all three task objectives: \par\nobreak
{\small
\begin{align}
    & \mathcal{L}_{plau} = -log(P_{plau}^{y_p}) \\
    & \mathcal{L}_{confl} = \begin{cases}  -log(P_{confl}^{y_c}) & \text{if $y_p$ = 0} \\
    0 & otherwise 
    \end{cases} \\
    & \mathcal{L}_{att} = -log(P(S^{p}|\phi^{p}, \psi^{p}))-log(P(S^{e}|\phi^{e}, \psi^{e})) \\ 
    & \mathcal{L} = \mathcal{L}_{plau} + \mathcal{L}_{confl} + \frac{1}{B} \sum^{b=0}_B \mathcal{L}_{att}^b
\end{align}
}
where $y_p$=0 if the story is not plausible and $y_p$=1 if the story is plausible, and $y_c$ denotes the conflict sentence pair index. Note that in our setup, each entity can produce a prediction for conflict sentence pair and story plausibility. At inference time, we take the average of all entities' logits to get the final predictions for these two objectives.  

\begin{table*}[!ht]
    \centering
    \caption{ProPara test set results. Modeling: E=entity, T=timestep-specific, GC=global context, GO=global outputs. 
    }
    \label{tab:propara_res}
    \resizebox{\textwidth}{!}{
    \begin{tabular}{l|rrrr|rrrrr|rrr}
    \toprule
        & \multicolumn{4}{c|}{Modeling} & \multicolumn{5}{c|}{Sentence-level} & \multicolumn{3}{c}{Document-level} \\
       \hline
       Model & E & T & GC & GO & Cat1 & Cat2 & Cat3 & Macro$^{avg}$ & Micro$^{avg}$ & P & R & F1 \\
       \hline
       ProLocal \cite{dalvi-etal-2018-tracking} & Y & Y & N & N & 62.7 & 30.5 & 10.4 & 34.5 & 34.0 & \bf 81.7 & 36.8 & 50.7 \\
       ProGlobal \cite{dalvi-etal-2018-tracking} & Y & Y & Y & N & 63.0 & 36.4 & 35.9 & 45.1 & 45.4 & 61.7 & 48.8 & 51.9 \\
       ProStruct \cite{tandon-etal-2018-reasoning} & Y & Y & N & Y & - & - & - & - & - & 74.3 & 43.0 & 54.5 \\
       KG-MRC \cite{das2018building} & N & Y & N & N & 62.9 & 40.0 & 38.2 & 47.0 & 46.6 & 64.5 & 50.7 & 56.8 \\
       NCET (\citeauthor{gupta-durrett-2019-tracking}) & N & N & Y & Y & 73.7 & 47.1 & 41.0 & 53.9 & 54.0 & 67.1 & 58.5 & 62.5 \\
       IEN \cite{tang-etal-2020-understanding-procedural} & N & N & Y & Y & 71.8 & 47.6 & 40.5 & 53.3 & 53.0 & 69.8 & 56.3 & 62.3 \\
       DynaPro \cite{DBLP:journals/corr/abs-2003-13878} & Y & Y & N & N & 72.4 & 49.3 & 44.5 & 55.4 & 55.5 & 75.2 & 58.0 & 65.5 \\
       TSLM (\citeyear{rajaby-faghihi-kordjamshidi-2021-time}) & Y & Y & Y & N & 78.8 & 56.8 & 40.9 & 58.8 & 58.4 & 68.4 & 68.9 & 68.6 \\
       KOALA \cite{zhang2021koala} & N & N & Y & Y & 78.5 & 53.3 & 41.3 & 57.7 & 57.5 & 77.7 & 64.4 & 70.4 \\
       \hline
       CGLI (Ours) & Y & Y & Y & Y & 80.3 & 60.5 & \bf 48.3 & \bf 63.0 & \bf 62.7 & 74.9 & \bf 70.0 & 72.4 \\
       CGLI (Ours) \small + Data Augmentation & Y & Y & Y & Y & \bf 80.8 & \bf 60.7 & 46.8 & 62.8 & 62.4 & 75.7 & \bf 70.0 & \bf 72.7 \\
       \bottomrule
    \end{tabular}
    }

\end{table*}

\section{Experimental Setup}
\label{sec:exp}
\textbf{Benchmarks.}
We evaluate procedural understanding on \texttt{ProPara}~\cite{mishra2018tracking}\footnote{\texttt{ProPara} is covered under Apache 2.0 License.}, which contains 488 human-written paragraphs from the natural science domain. The paragraphs are densely annotated by crowd workers, i.e., for every entity, its existence and location are annotated for every step. Additional 871 unannotated paragraphs are also provided by ProPara. We use these for data augmentation. 

We test story understanding on
\texttt{TRIP} \cite{storks-etal-2021-tiered-reasoning}, which contains crowdsourced plausible and implausible story pairs. In each pair, the plausible story label and the conflicting sentence pair label in implausible story are annotated. TRIP annotates 20 attributes with predefined set of possible values. The annotations are given for all entities at every timestep of the two stories.

We provide datasets splits details in \autoref{tab:stats}. For TRIP, we only report the unique story statistics in \autoref{tab:stats}. 
Note that \citet{storks-etal-2021-tiered-reasoning} have up-sampled some of the plausible stories to match the number of implausible stories.


\begin{table}[]
    \centering
    \caption{Statistics of the datasets.}
    \label{tab:stats}
    \begin{tabular}{l|lrrr}
    \toprule
        Dataset & Stats & Train & Dev & Test \\
       \hline
        & \#Paragraphs & 391 & 43 & 54 \\
       ProPara & \#Ents/Para & 3.8 & 4.1 & 4.4 \\
       & \#Sents/Para & 6.7 & 6.7 & 6.9 \\
       \midrule 
       & \#Paragraphs & 1169 & 474 & 504 \\
       TRIP & \#Ents/Para & 7.0 & 8.1 & 8.3 \\
       & \#Sents/Para & 5.1 & 5.0 & 5.1 \\
       \bottomrule
    \end{tabular}
\end{table}

\noindent \textbf{Evaluation metrics.}
\label{subsec:eval}
Following previous work, we report both sentence-level metrics\footnote{\url{https://github.com/allenai/propara/tree/master/propara/evaluation}} and document-level metrics\footnote{\url{https://github.com/allenai/aristo-leaderboard/tree/master/propara}} on ProPara. Sentence-level evaluation computes accuracy over three questions: whether the entity created (moved/destroyed) in the process (\textit{Cat1}), and if so, when (\textit{Cat2}) and where (\textit{Cat3}).\footnote{Cat2 and Cat3 only apply to entities that satisfy Cat1.} Document-level metrics compute F1 scores of the identified inputs (entities that exist before the process begins and are destroyed in the process), outputs (entities that do not exist before but exist after the process), conversions (instances where some entities are converted to other entities), and moves (location changes of entities).


For TRIP, we follow the original work and report the following metrics: \textit{accuracy} of classifying the plausible story, \textit{consistency} of finding the conflicting sentence pairs when the story classification is correct, and \textit{verifiability}, which evaluates the prediction of the entities' effects at $s_{c1}$ and the entities' preconditions at $s_{c2}$. We also report the average F1-score for preconditions and effects across the 20 attributes to better understand the model's procedural understanding ability. 

\noindent \textbf{Baselines.}
For ProPara, we directly report baseline results from the official leaderboard. 
For TRIP, we report the results from the best model released by \citet{storks-etal-2021-tiered-reasoning}. 

\noindent\textbf{Training details.} \label{sec:details}
For ProPara, we define two additional action types to represent the entity transitions, namely {Out-of-Create, Out-of-Destroy} similar to \cite{zhang2021koala}. Hence, the total size of the action space is six. For evaluation, these two types would be mapped to NONE transition, and they are defined to help the model differentiate the NONE types during training, i.e., if the entity has not being created or if it has been destroyed. 
To facilitate model's learning on location predictions, we initialized our model with a RoBERTa-Large \cite{liu2019roberta} model pretrained on SQuAD 2.0~\cite{rajpurkar-etal-2018-know}. We run our model five times with different random seeds and report the maximum scores in \autoref{tab:propara_res} and average scores with a 95\% confidence interval in \autoref{tab:abalation} and \autoref{tab:trip_res}. For TRIP, we directly initialize the model with RoBERTa-Large. On ProPara we train models for 20 epochs and 6 epochs with data augmentation to let the model receive the similar number of updates.  We train models for 10 epochs on TRIP. Except for training epochs, we use the same set of hyperparameters in all of our experiments: learning rate 1e-5, batch size 1, gradient accumulation 2. We used Transformers \cite{wolf-etal-2020-transformers} library \footnote{Covered under Apache 2.0 License.} for all of our experiments and all of our models have about 360M parameters.

\noindent\textbf{Computing infrastructure.}
We run our experiments on a single Nvidia A6000 GPU or a single Nvidia Titan RTX GPU. For ProPara, each experiment takes about 1.5 hours to finish. For TRIP, each experiment takes about 9 hours to finish. 


\section{Results and Analysis}


\subsection{Procedural text understanding}
\label{subsec:res}
CGLI significantly outperforms all previous baselines on ProPara, achieving state-of-the-art results (\autoref{tab:propara_res}). With data augmentation, our model achieves further improvement on document level. For each baseline, we indicate whether it considers entity-centric information (E), timestep-centric (T), global context (GC), and global output (GO). We note that models that adopt global output usually have much higher precision than recall on document level. On the other hand, TSLM is very good on recall, which is expected given its focus on entity and timestep input modeling.\footnote{This pattern may not always hold for other models due to other modeling differences, e.g., LSTM vs. BERT.}
CGLI is able to achieve both strong precision and recall, showing the benefit of global reasoning over both entity- and timestep-specific global inputs in a single model. 

\begin{figure*}[!t]
    \centering
    \includegraphics[width=0.85\linewidth]{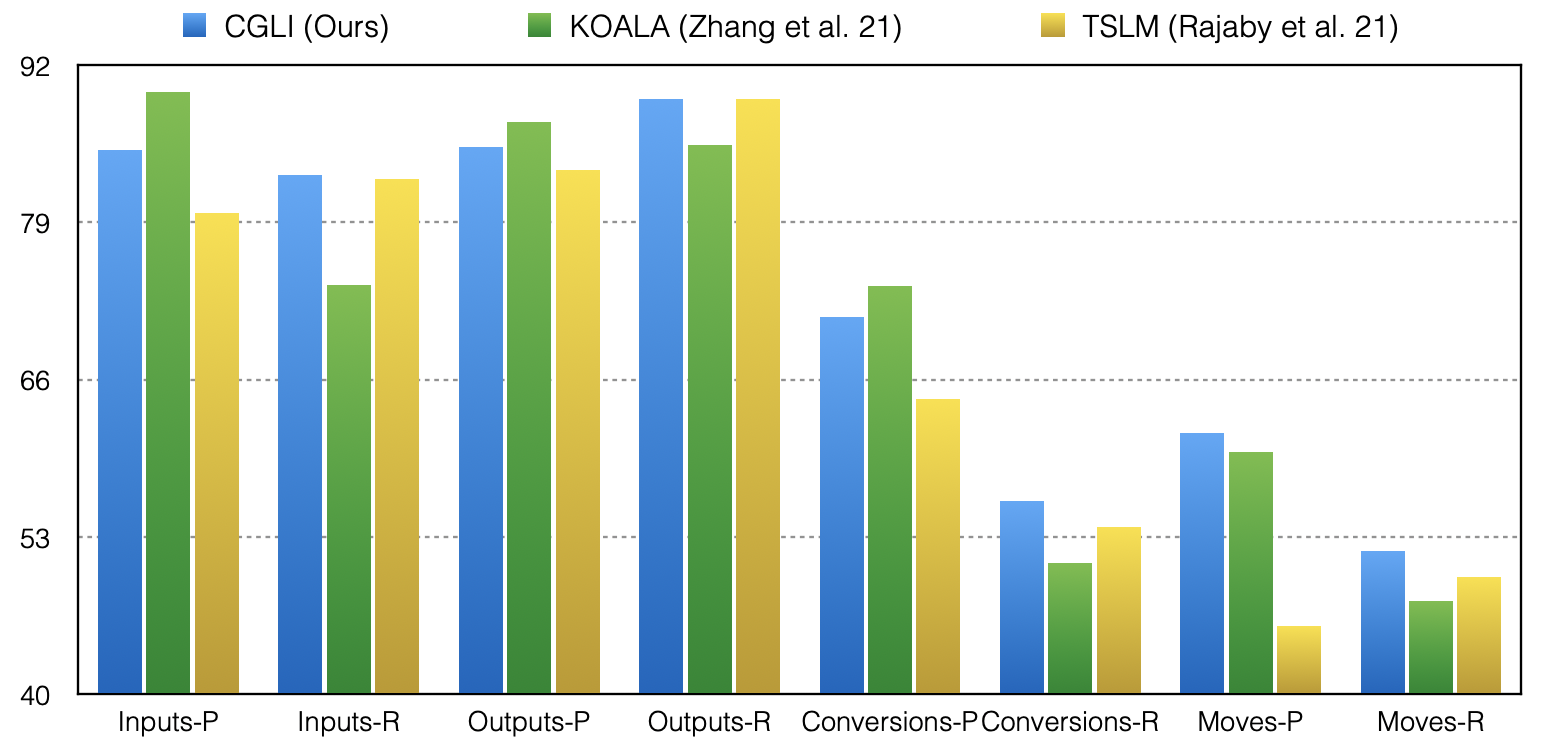}
    \caption{Document-level evaluation on ProPara test set, split by precision (P) and recall (R) per category (Inputs, Outputs, Conversions, Moves).}
    \label{fig:res}
\end{figure*}

We break down the results on ProPara by the document-level question types defined in \S\ref{sec:exp} and compare our best model with the best results reported by TSLM and KOALA. The precision and recall per question type are shown in \autoref{fig:res}. Consistent with the overall results, KOALA is particularly strong on precision for all types and TSLM is much better on recall. CGLI is able to maintain a balance between those two extremes and achieve overall better results. All three models perform similarly when predicting the inputs and the outputs of a procedure. Yet, CGLI achieves much higher performance on transitional questions regarding entity conversions and moves, which are notably harder to predict. These results suggest that the gains of CGLI over previous works are mostly due to hard-to-answer categories. 


\begin{table*}[!ht]
    \centering
    \small
    \caption{Document-level ablation results of proposed model components and modeling aspects on the ProPara.}
    \label{tab:abalation}
    \begin{tabular}{l|rrr|rrr}
    \toprule
        & \multicolumn{3}{c|}{Dev set} & \multicolumn{3}{c}{Test set}  \\
        \hline
        Model & P & R & F1  & P & R & F1 \\
       \hline
       CGLI \small + Data Augmentation & \small 78.5($\pm 1.7$) & \small 76.1($\pm 0.8$) & \small \bf 77.3($\pm 0.8$) & \small 75.2($\pm 1.1$) & \small 68.8($\pm 0.8$) & \small \bf 71.9($\pm 0.5$)\\
       CGLI & \small 77.3($\pm 1.5$) & \small 75.5($\pm 0.7$) & \small 76.4($\pm 1.0$) & \small 73.0($\pm 1.9$) & \small \bf 69.8($\pm 1.2$) & \small 71.3($\pm 0.9$)\\
       \small No SQuAD2.0 &  \small 76.5($\pm 1.3$) &  \small 75.4($\pm 0.9$) & \small 75.9($\pm 0.4$) & \small 72.5($\pm 2.7$) &  \small 68.0($\pm 1.3$) & \small 70.1($\pm 0.8$) \\
       \small No Prior & \small 75.6($\pm 0.8$) &  \small \bf 76.6($\pm 0.6$) & \small 76.1($\pm 0.3$) & \small 72.0($\pm 2.1$) &  \small 68.1($\pm 1.4$) & \small 70.0($\pm 1.3$) \\
       \hline 
       \small No GO &  \small 75.7($\pm 1.1$) &  \small 76.1($\pm 1.4$) & \small 75.9($\pm 0.5$) & \small 70.2($\pm 1.2$) &  \small 67.3($\pm 1.2$) & \small 68.7($\pm 0.8$) \\
       \small No GC & \small 75.5($\pm 1.3$) &  \small 73.2($\pm 1.0$) & \small 74.3($\pm 0.5$) & \small 73.2($\pm 2.2$) &  \small 66.7($\pm 0.6$) & \small 69.8($\pm 1.1$) \\
       \small No T & \small 82.3($\pm 0.7$) & \small 59.7($\pm 0.4$) & \small 69.2($\pm 0.3$) & \small 77.2($\pm 1.3$) & \small 54.3($\pm 1.0$) & \small 63.8($\pm 0.8$) \\
       \small No E & \small \bf 84.5($\pm 1.1$) & \small 48.6($\pm 0.3$) & \small 61.7($\pm 0.2$) & \small \bf 84.9($\pm 0.7$) & \small 40.8($\pm 0.5$) & \small 55.1($\pm 0.3$) \\
       \bottomrule
    \end{tabular}
\end{table*}

\begin{table*}[!t]
    \centering
    \caption{Results on the TRIP dataset. The F1 scores of last two columns are Macro averages of 20 attributes.}
    \small
    \label{tab:trip_res}
    \begin{tabular}{l|ccccc}
    \toprule
       Model & Accuracy & Consistency & Verifiability & Precondition F1 & Effect F1 \\
       \hline
       TRIP-RoBERTa \cite{storks-etal-2021-tiered-reasoning} & 73.2 & 19.1 & 9.1 & 51.3 & 49.3 \\
       \hline
       CGLI (Ours) & \small 93.4($\pm 1.5$) & \small 76.3($\pm 1.7$) & \small 24.8($\pm 1.6$) &  \small 70.8($\pm 1.8$) & \small 74.9($\pm 1.7$) \\
       CGLI (Ours) No CRF & \small \bf 94.1($\pm 0.7$) & \small \bf 77.3($\pm 1.0$) & \small \bf 28.0($\pm 2.5$) &  \small \bf 72.1($\pm 1.6$) & \small \bf 75.6($\pm 1.6$) \\ 
       \bottomrule
    \end{tabular}

\end{table*}

\subsection{Story understanding}

Our method outperforms the baseline method on the TRIP dataset by a very large margin on all metrics, especially on consistency where we observe nearly 400\% relative improvement over the baseline (\autoref{tab:trip_res}). This may seem surprising as both our model and the baseline use the same LM backbone. After further analysis of the baseline model, we notice three sub-optimal design decisions. First, the baseline detects conflicting sentence pairs via binary classification for every sentence, independently, without considering pairs of sentences. As a result, for 47.6\% of examples in TRIP test set, the baseline model predicted either less or more than two sentences as conflicting, thus getting a score of 0 on consistency. Second, the baseline uses the same encoded representations to directly model both story classification and conflicting pair detection objectives. Without using task-specific output projection layers, the model may be hard to optimize. Third, the baseline did not provide global input view to the model, i.e., each sentence is encoded independently. 

\begin{figure*}
    \centering
    \includegraphics[width=\textwidth]{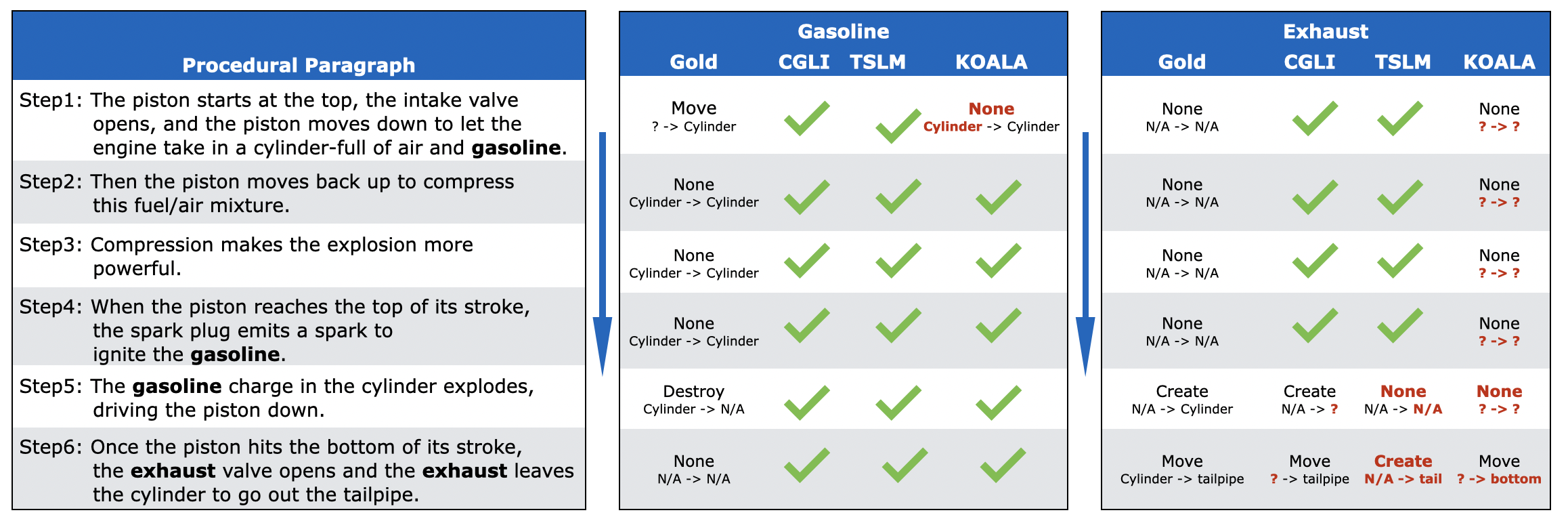}
    \caption{Example predictions on ProPara from three models for two entities. Red font indicate errors.}
    \label{fig:case}
\end{figure*}

\subsection{Ablation studies}

\noindent \textbf{Impact of modeling aspects} To understand the contribution of each of the four modeling aspects we identified for the procedural text understanding, we ablate each of them in CGLI. \\
\noindent \textbf{No GO} is done by removing the CRF layer and directly training the model with cross-entropy loss over the emission probability $\phi \in \mathbb{R}^{n \times a}$ defined in \S\ref{sec:model}. During inference, we predict the action at each timestep independently by taking the argmax over the emission probability instead of viterbi decoding. \textbf{No GC} is achieved by allowing the model model to access up to $t$ sentences at every timestep $t \in \{1, 2, 3, ... n\}$, i.e. the model has no access to future sentences. For \textbf{No T}, we remove the timestep embeddings such that each entity would have identical encoded context representations across timesteps. For \textbf{No E}, we no longer provide the pseudo question with the entity name in the input \S\ref{sec:model}, such that all entities in the same paragraph would have the same encoded context representations. \\
\noindent The results are shown in the bottom half of \autoref{tab:abalation}. Removing either T or E leads to drastic drop in the F1 score. This is not surprising because the model would have no clue how to distinguish different timesteps or different entities, respectively. We found that the model predict most of entity actions to be NONE, leading to extremely high precision and low recall. Removing GO also leads to a large drop in F1 score, which is actually similar to TSLM's performance, a model that lacks GO. This shows that modeling the global dependency is important for procedural understanding. Finally, removing GC also hurts the performance, which is also expected because location spans often only appear in future sentences, thus span extraction layer is at disadvantage in this setting.

\noindent \textbf{Impact of training data}
To understand the impact of the CGLI components, we ablate SQuAD2.0 pretraining by initializing the model with vanilla RoBERTa-Large model and we ablate prior initialization by randomly initializing the transition probabilities in the CRF layer. The results (upper half of \autoref{tab:abalation}) show that with data augmentation, CGLI achieves higher overall F1 scores on average and the gains are mostly from precision. Both pretraining on SQuAD2.0 and prior initialization have a positive impact on the CGLI performance.

As the continuity from effect to precondition states no longer holds on the TRIP story understanding task (cf. \S\ref{subsec:task1}), we investigate the impact of the CRF layers on modeling entity states. We remove the CRF layers for both effects and preconditions, and we directly train CGLI with regular classification objectives, hence entity states at each step are predicted independently (No GO). \autoref{tab:trip_res} shows that removing CRF improves performance. We hypothesize that this is caused by the implausible stories in the dataset. Since the entity states in the implausible story's conflicting sentences are inconsistent by nature, training the CRF to maximize their probabilities can be confusing for the model. To verify this, we train models with and without CRF on plausible stories only. In this case, the model is only trained to predict entities effects and preconditions. We found that the models have very similar F1 scores with or without CRF (preconditions 74.1 vs 73.7, effects 76.5 vs 76.6). Thus, we conclude that implausible stories are detrimental to CRF training. Moreover, as the effects of the previous step are not a precondition of the current step on TRIP, the outputs from previous steps can hardly contribute to the current prediction, thus CRF has a limited contribution even on the plausible stories.

\begin{table}[!t]
    \centering
    \caption{Error Examples on TRIP. The conflicting pairs are marked with *, and the entity of interest with \textit{italic}.}
    \label{tab:trip_error}
    
    \resizebox{\linewidth}{!}{
    \begin{tabular}{l}
    \toprule
      Ann washed her hair in the bathtub. \\
      Ann used the hair dryer to get ready to go out. \\
      Ann applied deodorant to her armpits. \\
      *\textit{Ann} put her pants on. \\
        - (Effects, is wet), Pred: False,  Gold: Irrelevant \\
      *Ann ironed her \textit{pants} before going out. \\
        - (Preconditions, is wet), Pred: True, Gold: Irrelevant \\
       \hline  
       *John forgot his \textit{notebook} at home. \\
        - (Effects, location), Pred: Moved, Gold: Irrelevant \\
       John sat at his desk. \\
       John opened up his book bag. \\
       * John took out his \textit{notebook}. \\
        - (Preconditions, location), \\
        - Pred: Picked up, Gold: Taken out of container \\
       John began writing down notes. \\
       
       \bottomrule
    \end{tabular}
}
\end{table}

\subsection{Case Studies}

We show an example of tracking states for two entities from ProPara with partial outputs from CGLI, TSLM, and KOALA in \autoref{fig:case}. For gasoline, our model and TSLM both got perfect predictions, but KOALA missed the action at step 1, thus predicting no moves across the process. For exhaust, the sentence in step 6 gives a strong signal for a movement, however, there is no mention of exhaust in the previous steps. Our model is able to infer that \textit{create} needs to come before \textit{move}, thus correctly predicting the actions in steps 5 and 6. However, since TSLM does not have the global output view, it cannot capture such transitions. For KOALA, although it is also able to predict the move and infer that the exhaust should exist before the move, it is unable to predict the create action. We note that for both entities, KOALA is more reluctant to predict actions compared to the other two models. These observations explain why KOALA achieves overall higher precision but lower recall. 

We show story reasoning examples from TRIP in \autoref{tab:trip_error}. Since the largest gap in the model performance is between consistency and verifiability, we select examples where our model successfully predicted conflicting sentences but failed to predict entity states. We see that the model still lacks common sense on certain concepts, e.g., forgetting something at home does not result in changing its location, and people usually iron their clothes after they are dry. We also note that some entity states might be hard to distinguish, e.g., the distinction between picking up something versus taking something out of a container only depends on previous location of the object, which might be hard for models to learn from data. These observations suggest that enhancing the model's commonsense reasoning ability is a promising future direction.




\section{Related Work}
Recent \textbf{procedural text understanding} benchmarks including ScoNe \cite{long-etal-2016-simpler}, bAbI \cite{weston2015aicomplete}, ProcessBank \cite{berant-etal-2014-modeling}, ProPara~\cite{mishra2018tracking}, Recipe ~\cite{Bosselut2017SimulatingAD}, and OpenPI \cite{tandon-etal-2020-dataset}  have inspired a series of methods. \citet{mishra2018tracking} propose ProLocal that encodes each step of a procedure separately and ProGlobal that encodes the full paragraph at every step. KG-MRC \cite{das2018building} builds a dynamic knowledge graph of entity and location mentions to communicate across time steps. DynaPro \cite{DBLP:journals/corr/abs-2003-13878} employs pre-trained LM to jointly predict entity attributes and their transitions. TSLM \cite{rajaby-faghihi-kordjamshidi-2021-time} formulates procedural understanding as a question answering task, and leverages models pretrained on SQuAD \cite{rajpurkar-etal-2016-squad} enhanced with a timestamp encoding. Although equipped with various ways to pass information across time steps, these methods still make local predictions thus they may compromise the global dependency of outputs. Another line of work focuses on jointly modeling the entity action sequence, aiming to ensure global structure and consistency. ProStruct \cite{tandon-etal-2018-reasoning} aims to find the globally optimal entity action sequence using beam search. \citet{gupta-durrett-2019-tracking} devise a structured neural architecture NCET, modeled with a CRF, which recurrently updates the hidden representation of each entity at each step. IEN \cite{tang-etal-2020-understanding-procedural} builds upon NCET and augments the entity-to-entity attention. KOALA~\cite{zhang2021koala} further enhances the NCET by pretraining on Wikipedia 
and ConceptNet~\cite{speer2017conceptnet}. The key shortcoming of these global methods is that they rely on entity mentions extracted from a single copy of encoded context shared by all entities and all steps, which limits their modeling capacity. Our proposed method stands out from all previous works by coalescing complementary granularities of procedural text modeling, by building specific and informative input representations while modeling output dependency. Concurrent to our work, \citet{shi2022lemon} proposed LEMON for language-based environment manipulation. Their focus on model pretraining is orthogonal to CGLI.

There are also numerous recent \textbf{story understanding} benchmarks \cite{mostafazadeh-etal-2016-corpus,qin-etal-2019-counterfactual,mostafazadeh-etal-2020-glucose}, and modeling methods \cite{qin-etal-2020-back,guan-etal-2020-knowledge,Gabriel2021ParagraphLevelCT,ghosal-etal-2021-stack}. The TRIP task \cite{storks-etal-2021-tiered-reasoning} integrates a procedural understanding component in story understanding to enable consistent and interpretable reasoning over narratives. To our knowledge, we are the first work to bridge the gap of modeling methods between procedural understanding and story comprehension. Other tasks that require reasoning over procedures exist, including defeasible reasoning \cite{rudinger-etal-2020-thinking,madaan-etal-2021-think},
abductive commonsense inference~\cite{ch2019abductive}, reasoning over preconditions  \cite{qasemi2021corequisite}, 
script reasoning \cite{zhang-etal-2020-reasoning,sakaguchi-etal-2021-proscript-partially} and multimodal script reasoning \cite{yang-etal-2021-visual,wu2021understanding}, are typically solved by specialized methods, without separately modeling procedural and causal links. We intend to apply CGLI on these tasks in the future to bridge this gap.


\section{Conclusions \& Future Work}
We proposed CGLI: a novel procedural understanding method that combined global and local information. Recognizing the key role of procedural understanding in downstream tasks, we also integrated CGLI in a story understanding framework. Our experiments showed the benefit of our coalesced method, with the global views providing optimal precision, while the local view boosting its recall, ultimately achieving new state-of-the-art results. We demonstrated that CGLI can help with classifying stories, and identifying the conflicting sentence for inconsistent stories. Future work should investigate how to enhance the commonsense ability of our procedural understanding model, e.g., by injecting commonsense knowledge during finetuning \cite{chen2018incorporating,ma-etal-2019-towards} or by pretraining on commonsense knowledge bases \cite{guan-etal-2020-knowledge,ilievski2021story,ma2020knowledgedriven},
and how to apply procedural understanding to other downstream tasks, such as dialogue modelling \cite{zhou-etal-2021-probing-commonsense} and planning \cite{2020}. Also, it's worth exploring the lightweight-tuning methods \cite{ma-etal-2021-exploring,vu-etal-2022-spot} to enhance the model's generalization and reduce computation cost. 

\section*{Acknowledgements}
We would like to thank Yonatan Bisk, Aman Madaan, and Ruohong Zhang for helpful discussions and the anonymous reviewers for their valuable suggestions on this paper. Some datasets and models are used by this work, despite their not having specified licenses in their code repositories---for these, we followed previous works and only used them for pure research purposes. 

\clearpage

\bibliography{anthology,custom}
\bibliographystyle{acl_natbib}

\clearpage
\appendix

\end{document}